\pdfoutput=1

\documentclass[11pt]{article}

\usepackage[final]{acl}

\usepackage{times}
\usepackage{latexsym}
\usepackage{subcaption}
\usepackage{caption}
\usepackage{amsmath} 
\usepackage[T1]{fontenc}

\usepackage[utf8]{inputenc}
\usepackage{booktabs} 
\usepackage{microtype}

\usepackage{inconsolata}
\usepackage{multirow}
\usepackage{multicol}

\newcommand{\tablestyle}[2]{\setlength{\tabcolsep}{#1}\renewcommand{\arraystretch}{#2}\centering\footnotesize}

\newcommand{\tableCellHeight}{1}
\newcommand{\tabstyle}[1]{
  \setlength{\tabcolsep}{#1}
  \renewcommand{\arraystretch}{\tableCellHeight}
  \centering
  \small
}

\usepackage{color}
\usepackage{enumerate}
\usepackage{color}
\usepackage{graphicx}
\usepackage{amssymb}
\usepackage{pifont}
\usepackage{enumerate}
\usepackage{graphicx}

\title{Vision-Language Model Fine-Tuning via Simple Parameter-Efficient Modification}

\author{
  \textbf{Ming Li \textsuperscript{1}},
  \textbf{Jike Zhong\textsuperscript{2}},
  \textbf{Chenxin Li\textsuperscript{4}},
  \textbf{Liuzhuozheng Li\textsuperscript{1}},
\\
  \textbf{Nie Lin\textsuperscript{1}},
  \textbf{Masashi Sugiyama\textsuperscript{3,1}}
\\
  \textsuperscript{1}The University of Tokyo,
  \textsuperscript{2}University of Southern California,\\
  \textsuperscript{3}RIKEN Center for Advanced Intelligence Project,\\
  \textsuperscript{4}	The Chinese University of Hong Kong
\\
  \small{
li-ming948@g.ecc.u-tokyo.ac.jp \quad 
jikezhon@usc.edu}\\
\small{chenxinli@stu.xmu.edu.cn \quad  l.li@ms.k.u-tokyo.ac.jp}\\
\small{nielin@iis.u-tokyo.ac.jp\quad sugi@k.u-tokyo.ac.jp}
}

\begin{document}
\maketitle
\begin{abstract}

Recent advances in fine-tuning Vision-Language Models (VLMs) have witnessed the success of prompt tuning and adapter tuning, while the classic model fine-tuning on inherent parameters seems to be overlooked. It is believed that fine-tuning the parameters of VLMs with few-shot samples corrupts the pre-trained knowledge since fine-tuning the CLIP model even degrades performance. In this paper, we revisit this viewpoint, and propose a new perspective: fine-tuning the specific parameters instead of all will uncover the power of classic model fine-tuning on VLMs. Through our meticulous study, we propose ClipFit, a simple yet effective method to fine-tune CLIP without introducing any overhead of extra parameters. We demonstrate that by only fine-tuning the specific bias terms and normalization layers, ClipFit can improve the performance of zero-shot CLIP by 7.27\% average harmonic mean accuracy. Lastly, to understand how fine-tuning in CLIPFit affects the pre-trained models, we conducted extensive experimental analyses w.r.t. changes in internal parameters and representations. We found that low-level text bias layers and the first layer normalization layer change much more than other layers.
The code is available at \url{https://github.com/minglllli/CLIPFit}. 
\end{abstract}

\section{Introduction}
\label{sec:intro}

\begin{figure*}[t]
\begin{center}
    \includegraphics[width=0.85\linewidth]{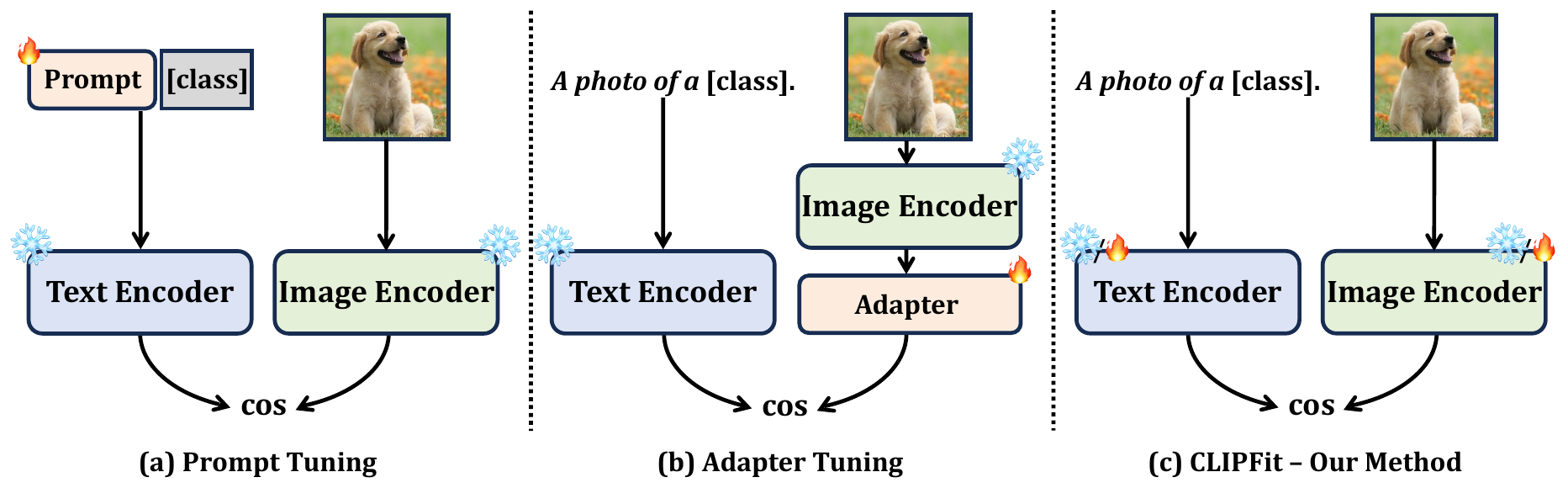}
\end{center}
\vspace{-0.5em}
\caption{Comparison of (a) prompt tuning methods, (b) adapter tuning methods, and (c) our proposed CLIPFit method. Prompt tuning methods introduce a set of learnable external parameters as input to learn task-specific knowledge. Adapter tuning methods introduce extra learnable networks following the image encoder to learn task-specific features. Unlike these two methods, our CLIPFit does not introduce external parameters and fine-tunes only a small portion of the CLIP model.
} 
\label{fig:compa}
\vspace{-0.7em}
\end{figure*}

\indent Large pre-trained Visual-Language Models (VLMs) have been developed a lot in recent years. 
For example, CLIP \cite{radford2021learning} and 
ALIG \cite{jia2021scaling} demonstrated remarkable performance for various tasks, e.g., image recognition in a zero-shot fashion.
To further improve the performance on the specific downstream tasks, prompt tuning \cite{ lester2021power,yao2023visual,zhu2023prompt,zhou2022conditional} and adapter tuning \cite{gao2023clip,zhang2021tip} methods have been proposed. 
As shown in Fig. \ref{fig:compa}, prompt tuning methods proposed to introduce a set of learnable prompt vectors as the input of the text encoder while adapter tuning approaches adopted an additional bottleneck layer to learn new features. During the fine-tuning procedure, both of these two strategies keep CLIP's parameters fixed. The performance of prompt tuning and adapter tuning methods are superior on various tasks \cite{zhou2022learning,gao2023clip}, so research on fine-tuning the inherent parameters of VLMs has been barely touched.

For language models, fully fine-tuning with downstream data can achieve promising results \cite{zaken2021bitfit,liu2022p}. Moreover, recent works in language model fine-tuning (e.g., BitFit \cite{zaken2021bitfit}) have demonstrated that, without introducing any external parameters, 
fine-tuning only the bias terms in a pre-trained model can perform competitively on downstream tasks compared with fine-tuning the entire model. For VLMs, however, it is believed that fine-tuning the parameters of VLMs corrupts the inherent pre-trained knowledge as fully fine-tuning degrades performance \cite{zhou2022learning}. In this paper, we revisit this viewpoint and ask if,  without introducing any external parameters, fine-tuning the inherent parameters of VLMs can achieve competitive performance compared with prompt tuning. 

We start with directly applying BitFit to fine-tuning the CLIP model. We explore two strategies: (i) applying BitFit to the text encoder alone, and (ii) applying BitFit to both the text and image encoder. We found that both two strategies can acquire task-specific knowledge but their performance to unseen class data can be poor (more discussed in Sec. \ref{sec:abla}), implying that directly fine-tuning the bias terms of a text or image encoder may harm the model's generalization ability. These findings motivate us to develop more effective and efficient fine-tuning techniques for VLMs.

In light of this, we propose CLIPFit, a simple yet effective method for efficiently fine-tuning VLMs. CLIPFit is orthogonal to previous prompt tuning and adapter tuning methods, as shown in Fig. \ref{fig:compa} (c). For the text encoder, instead of fine-tuning all the bias terms, CLIPFit proposes to tune only the bias terms of projection linear layers in feed-forward networks (FFNs). Fine-tuning only the bias terms of projection linear layers in FFNs
will reduce the number of training parameters compared with fine-tuning all the bias terms. Moreover, empirically, we discovered that our bias term tuning strategy can generalize better than BitFit \cite{zaken2021bitfit}, as shown in Sec. \ref{sec:abla}. For the image encoder, as mentioned before, it may harm the model's performance if directly applying BitFit. In the image encoder, layer normalization (LayerNorm) \cite{ba2016layer} aims to normalize the distributions of intermediate layers. Since the distributions of pre-training and downstream data might be divergent, pre-trained LayerNorm might lead to sub-optimal performance for downstream data inference. Therefore, CLIPFit proposes to further update only the parameters of the image encoder's LayerNorm. Updating LayerNorm can yield a better image encoder for downstream data. Lastly, previous studies \cite{yao2023visual} have shown that generic pre-trained knowledge is easily forgotten in the fine-tuning stage. Therefore, we explored two different regularization strategies for alleviating forgetting: (i) using the knowledge distillation (KD) loss \cite{hinton2015distilling} to guide CLIPFit to learn from the zero-shot CLIP; (ii) using the mean squared error (MSE) loss in bias terms to penalize changes in text encoder. We empirically found that both two strategies can alleviate forgetting problems and the KD loss performs better, thus we used the KD loss as the final solution for CLIPFit.

Fine-tuning is an empirical and black-box process. So, understanding how fine-tuning affects the
pre-trained models is important for uncovering the black-box fine-tuning process. Previous works \cite{zhou2022closer,de2020s,merchant2020happens} explored this for language models fine-tuning. However, very little work explored the internal black-box fine-tuning process for VLMs. In this paper, we conducted an initial exploration to analyze VLM fine-tuning process of CLIPFit, focusing on changes in internal parameters and representations. We found that for bias terms in the FNN of the text encoder, as the number of layers increases, the change in bias decreases, which means that during the fine-tuning process, low-level features in the text encoder change more than high-level features.  For LayerNorm in the image encoder, we found that the first layer (patch embedding) changes much more than other layers. Experimentally, we showed that more changed layers play a more important role in adapting downstream knowledge than less changed layers. Moreover, we explored how KD loss affects the fine-tuning process for alleviating forgetting. We found that KD loss will reduce the changes for the more-changed low-level bias terms and enhance changes in less-changed high-level layers, which implies that penalizing changes for low-level bias terms is important for avoiding overfitting. Lastly, we found that tuning LayerNorm will form a better image feature space compared with zero-shot CLIP.

We conducted extensive experiments on 11 datasets in 4 different settings to show the effectiveness of the proposed CLIPFit. Overall, our main contributions can be summarized as follows\vspace{-0.5em}:
\begin{itemize}
  \item We propose a CLIPFit method for efficiently fine-tuning the CLIP model to uncover the power of classic model fine-tuning on VLMs. Unlike existing prompt tuning or adapter tuning methods, CLIPFit does not introduce any external parameters and only fine-tunes a small specific subset of CLIP's inherent parameters.
\item 
To analyze how CLIPFit affects the
pre-trained models,  we conducted extensive analyses during the fine-tuning process, focusing on the changes in parameters and representations. These analyses help us better understand the black-box fine-tuning process.
\item  We conducted extensive experiments on 11 datasets. Results show that CLIPFit brings a 7.33\% improvement in harmonic mean accuracy compared with zero-shot CLIP on the 16-shot base-to-new setting,  demonstrating that
CLIPFit is a promising alternative to prompt tuning and adapter tuning.
\end{itemize}

\section{Related Works}
\label{works}
\textbf{Visual-Language Models (VLMs).}
With large-scale  available web-crawled
image-text pairs \cite{schuhmann2022laion}, pre-training VLMs have been developed fast in recent years \cite{xu2021videoclip,radford2021learning, jia2021scaling,wang2022medclip} and achieved
remarkable zero-shot performance in the downstream
tasks, e.g., image classification. Despite the remarkable transfer ability, the potential of VLMs can be further stimulated by fine-tuning it with few-shot downstream data \cite{song2022clip,zhang2021tip,shen2021much,wang2022learning,wang2022dualprompt,chen-etal-2023-mclip}. 

\textbf{Parameter-efficient Fine-tuning (PEFT) on VLMs.} There are mainly two categories of VLM parameter-efficient fine-tuning methods: prompt tuning \cite{zhou2022learning,zhou2022conditional,chen2022prompt,yao2023visual,zhu2023prompt,zhang2023prompt,khattak2023maple} and adapter tuning \cite{gao2023clip,zhang2021tip}. Prompt tuning methods for VLMs introduced a few learnable parameters (prompts) as input, which were inspired by language prompt tuning \cite{lester2021power}. Adapter tuning methods set an additional bottleneck layer following the text or image encoder to learn better features by a residual way. Both prompt tuning and adapter tuning methods boost CLIP's performance, so research on fine-tuning the inherent parameters of CLIP seems to be overlooked. To explore classic model fine-tuning on VLMs, our CLIPFit proposes to fine-tune CLIP by modifying a small portion of the CLIP model's inherent parameters without introducing any external learnable parameters.

\textbf{PEFT on Large Language Models.} Fully fine-tuning language models \cite{radford2018improving, devlin2018bert} can achieve promising results but is expensive. To efficiently fine-tune pre-trained language models, a lot of approaches have sought to fine-tune only a small number of parameters. For example, adapter methods \cite{bapna2019simple,houlsby2019parameter,pfeiffer2020mad} and prompt tuning methods \cite{liu2023pre, lester2021power,brown2020language,gao2020making} introduce a set of learnable external parameters for adaptation to downstream tasks. Recently, BitFit \cite{zaken2021bitfit} demonstrated that, without introducing any new parameters, fine-tuning only the bias terms in pre-trained language models can perform competitively compared with fully fine-tuning. 
However, BitFit is designed for LLM fine-tuning, and our experiments in Sec. \ref{exp} shows that directly applying BitFit to VLM fine-tuning may harm the model's generalization ability. Thus, our CLIPFit proposes to only fine-tune the LayerNorm of image encoder motivated by distribution shift.
Our method is different to BitFit 
Moreover, to understand how fine-tuning affects pre-trained models, various works \cite{zhou2022closer,mosbach2020interplay,de2020s,merchant2020happens} have explored this with LLM fine-tuning. However, very little work was attempted on the VLM side. In this paper, we attempt to bridge this gap by conducting an initial exploration to analyze the fine-tuning process in CLIPFit for VLMs, focusing 
on changes in internal parameters and representations. 

\begin{figure*}[t]
\begin{center}
    \includegraphics[width=0.9\linewidth]{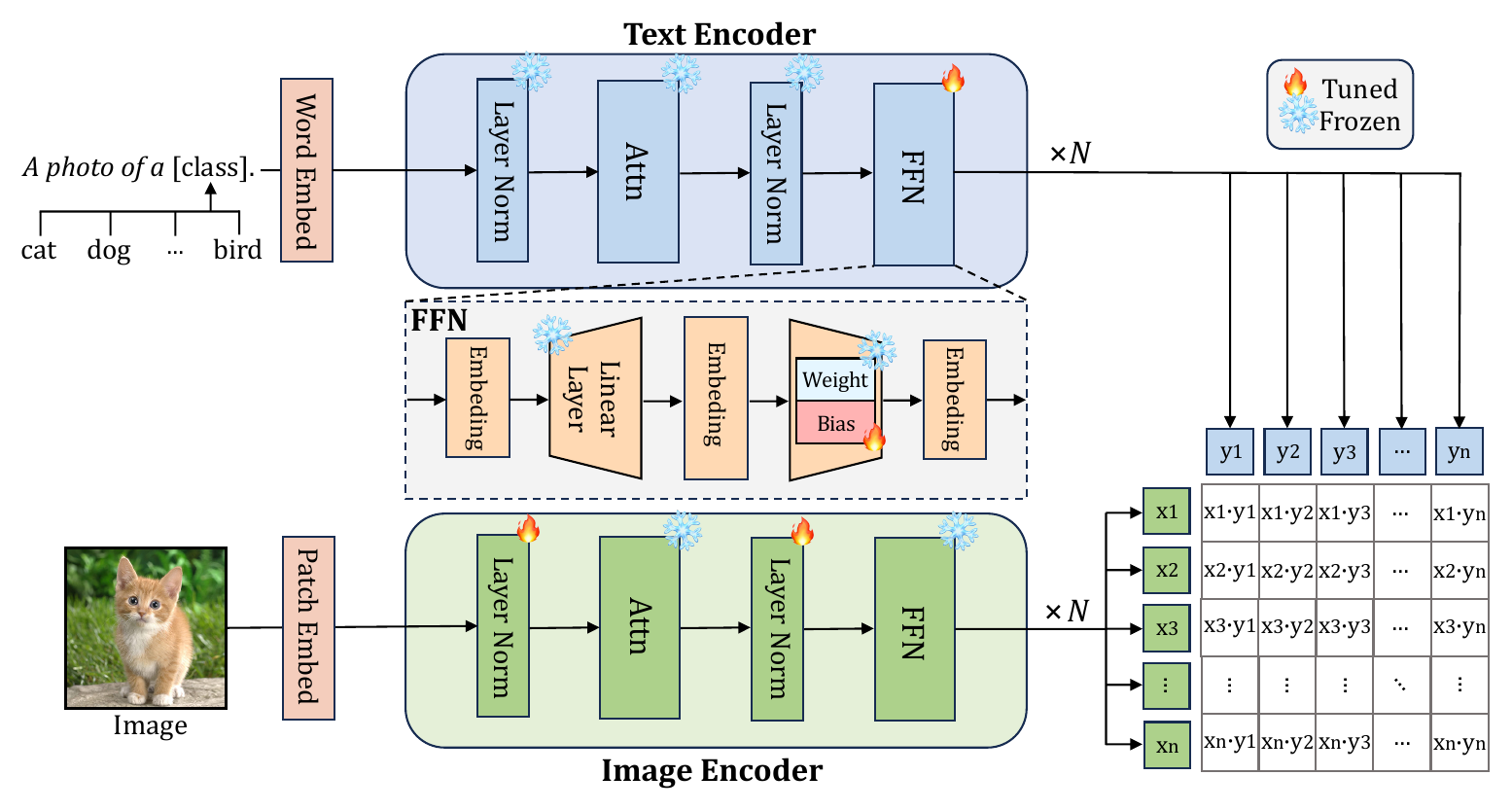}
\end{center}
\vspace{-0.1em}
\caption{An overview of our CLIPFit. Unlike existing prompt tuning methods or adapter tuning methods, CLIPFit does not introduce any external parameters and fine-tunes specific inherent parameters of CLIP. For the text encoder, as shown in the upper part of the figure, CLIPFit fine-tunes only the bias terms of projection linear layers in feed-forward networks. For the image encoder, as shown in the lower part of the figure, CLIPFit updates LayerNorm.
} 
\label{fig:framework}
\vspace{-0.5em}
\end{figure*}
\section{Methodolgy}
In this section, we introduce CLIPFit. We first briefly review CLIP and then illustrate CLIPFit.
\label{method}

\subsection{Review of CLIP} We first briefly review CLIP \cite{radford2021learning}. During pre-training, CLIP aims to align image features and text features in the joint embedding space to capture the relationship between images and texts. Let $D=\{(\boldsymbol{x}_i,\boldsymbol{t}_i)\}_{i=1}^b$ be the sampled batch, where $\boldsymbol{x}_i$ is the input image, $\boldsymbol{t}_i$ is the input text and $b$ is the batch size. A CLIP model is comprised of two types of encoders: visual encoder $E_\mathrm{I}(\cdot, \boldsymbol{\theta}_\mathrm{I} )$ and text encoder $E_\mathrm{T} (\cdot, \boldsymbol{\theta}_\mathrm{T})$. The visual encoder encodes image $\boldsymbol{x}_i$ into $\boldsymbol{f}_i$ and text $\boldsymbol{t}_i$ into $\boldsymbol{g}_i$, i.e., 
\begin{equation}
    \boldsymbol{f}_i = E_\mathrm{I}(\boldsymbol{x}_i, \boldsymbol{\theta}_\mathrm{I}), \indent \boldsymbol{g}_j=E_\mathrm{T}(\boldsymbol{t}_i, \boldsymbol{\theta}_\mathrm{T}).
\end{equation}
Then, a contrastive learning loss is applied to them for alignment. 

After pre-training, CLIP can perform zero-shot image recognition by comparing the image features with class weights $\{\boldsymbol{w}_i\}_{i=1}^K$, where $K$ is the number of classes. The class weight $\boldsymbol{w}_i$ is generated by text encoder $E_T(\cdot, \boldsymbol{\theta}_T)$ which takes the class descriptions (prompts) as input. These prompts usually take the form ``a photo of a [CLASS].'', where the class token will be replaced by the $i$-th class name (e.g., cat) for weight $\boldsymbol{w}_i$. Formally, for an image feature $\boldsymbol{f}$, the probability that it belongs to class $i$ is calculated by
\begin{equation}
    p(y=i \mid \boldsymbol{x})=\frac{\exp \left(\cos \left(\boldsymbol{w}_{{i}}, \boldsymbol{f}\right) / \tau\right)}{\sum_{j=1}^K \exp \left(\cos \left(\boldsymbol{w}_{{j}}, \boldsymbol{f}\right) / \tau\right)},
\end{equation}
where $\tau$ is a temperature parameter learned by CLIP during pre-training and $\cos(\cdot, \cdot)$ denotes the cosine similarity function.

\subsection{CLIPFit}
\label{clipfit}
The overall pipeline of the proposed CLIPFit is shown in Fig. \ref{fig:framework}. Without introducing any external parameters, CLIPFit involves fine-tuning only the bias terms of projection linear layers in FNNs of the text encoder and updating LayerNorm \cite{ba2016layer} in the image encoder. 

\textbf{Text Encoder.} For the text encoder, instead of fine-tuning all bias terms, CLIPFit fine-tunes only the bias terms of projection linear layers (i.e., second layers) in the FFNs of the text encoder. Fine-tuning only part of bias terms will reduce the number of training parameters compared with fine-tuning all bias terms. Moreover, Sec. \ref{sec:abla} will empirically show that our bias tuning method can achieve better performance compared with fine-tuning all bias terms \cite{zaken2021bitfit}.

\textbf{Image Encoder.} As mentioned in Sec. \ref{sec:intro}, directly applying BitFit \cite{zaken2021bitfit} to the image encoder may cause a negative impact on the model's performance. Instead of fine-tuning the bias terms of the image encoder, CLIPFit proposes to fine-tune LayerNorm. In LayerNorm, the two learniable parameters gain $\boldsymbol{g}$ and bias $\boldsymbol{b}$ are applied for affine transformation on normalized input vectors $\boldsymbol{x}$ for re-centering and re-scaling, which are expected to enhance the expressive power by re-shaping the distribution \cite{ba2016layer}. Different data distributions should produce different gains and biases in LayerNorm for distribution re-shaping during the training process. Prior work has also found norm layers to be an influential component in the fine-tuning process \cite{giannou2023expressivepowertuningnormalization, qi2022parameterefficienttuninglayernormalization, tu2023holistictransfernondisruptivefinetuning, zhong2024makingbatchnormalizationgreat}. So, if shifted gains and biases in LayerNorm are applied during inference, it may lead to a sub-optimal solution. Therefore, CLIPFit proposes to fine-tune LayerNorm in the image encoder.
 
\textbf{Loss function.} Previous works \cite{yao2023visual,xuhong2018explicit} have verified that during the fine-tuning stage, generic pre-trained knowledge is easily forgotten. Therefore, we explore two different strategies for alleviating such forgetting. The first one is to use the knowledge distillation \cite{hinton2015distilling,yao2023visual} loss to guide CLIPFit to learn from the original zero-shot CLIP. Let $\{\boldsymbol{w}_i^\mathrm{clip}\}_{i=1}^K$ and $\{\boldsymbol{w}_{i}\}_{i=1}^K$be the text features from original CLIP and text features from CLIPFit. The training loss and KD loss of CLIPFit are defined by
\begin{equation}
    \mathcal{L}=\mathcal{L}_{\mathrm{ce}}+\beta \mathcal{L}_{\mathrm{k g}},
\end{equation}
\begin{equation}
    \mathcal{L}_\mathrm{k g} = \frac{1}{K}\sum_{i=1}^{K}\cos(\boldsymbol{w}_i^{\mathrm{clip}},\boldsymbol{w}_i),
\end{equation}
where $\mathcal{L}_\mathrm{c e}$ is the cross entropy loss for classification \cite{zhou2022learning, zhou2022conditional} and $\beta$ is a hyperparameter.

The second strategy is using the MSE loss in bias terms
to penalize changes in the text encoder. Let $\{\boldsymbol{b}_i^\mathrm{clip}\}_{i=1}^L$ and $\{\boldsymbol{b}_i\}_{i=1}^L$ be the unfixed text bias terms from pre-trained CLIP and unfixed text bias terms from CLIPFit, where $L$ is the number of unfixed bias layers. The MSE loss is defined as\vspace{-0.5em}
\begin{equation}
    \mathcal{L}_\mathrm{m s e} = \frac{1}{L}\sum_{i=1}^{L}||\boldsymbol{b}_i^\mathrm{clip}-\boldsymbol{b}_i||^2.\vspace{-0.6em}
\end{equation}
We found that both strategies can alleviate the forgetting problems and the KD loss performs better (as discussed in Sec. \ref{sec:anan}), thus we adopted the KD loss as the final solution for CLIPFit.
\section{Experiments}
\label{exp}
In this section, we show and discuss the experimental results. To evaluate the effectiveness of our proposed method, we conducted extensive experiments and analyses on 11 datasets. 
\subsection{Experimental Setup}
\textbf{Datasets.}  Following CoOp, we conducted extensive experiments on 11 public classification benchmark datasets to evaluate CLIPFit. The datasets are  ImageNet \cite{deng2009imagenet}, Caltech101 \cite{fei2004learning},
OxfordPets \cite{parkhi2012cats}, StanfordCars \cite{krause20133d}, Flowers102 \cite{nilsback2008automated},
Food101 \cite{bossard2014food}, FGVCAircraft \cite{maji2013fine}, SUN397 \cite{xiao2010sun}, DTD \cite{cimpoi2014describing},
EuroSAT \cite{helber2019eurosat}, and UCF101 \cite{soomro2012ucf101}.
 \begin{table}[t!]
\centering
\tablestyle{-12pt}{1.1}
\addtolength{\tabcolsep}{+14pt}
\resizebox{\columnwidth}{!}{%
\begin{tabular}{lc|c c c c c c| c}
\toprule
{Dataset}& &    {CLIP} &  {CoOp} & {CoCoOp} &  {Adapter} & {KgCoOp}&
{MaPLe} & 
{CLIPFit} \\
\midrule
\multirow{3}{*}{\shortstack[l]{Average}}       & Base      & 69.34  & 82.69   & 80.47  &82.23   &80.73 & 82.28        & \textbf{83.72}  \\
                               & New     & 74.22  & 63.22   & 71.69  & 70.61    &73.60& \textbf{75.14}          &  74.84 \\
                               & HM        & 71.70  & 71.66   & 75.83 &	75.98     &77.00& 78.55        & 		\textbf{79.03} \\
\midrule
\multirow{3}{*}{ImageNet}      & Base   & 72.43  & 76.47   &  75.98 &  76.13      &75.83 & \textbf{76.66}         &76.2 \\ 
                               & New    & 68.14  & 67.88   &  70.43 & 67.17      & 69.96&\textbf{70.54}       & 70.17   \\ 
                               & HM          & 70.22  & 71.92   &  73.10 & 71.37   &72.78& \textbf{73.47 }    &73.06 \\
\midrule
\multirow{3}{*}{Caltech101}    & Base   & 96.84  & 98.00   &  97.96 & 97.40  &97.72   & 97.74          & \textbf{98.3} \\
                               & New     & 94.00  & 89.81   &  93.81 & 93.23   &93.70 & \textbf{94.36}  & 93.7\\
                               & HM          & 95.40  & 93.73   &  95.84 &  95.51  &\textbf{96.03} &{96.02} & 95.94 \\
\midrule
\multirow{3}{*}{OxfordPets}    & Base     & 91.17  & 93.67   & 95.20 & 94.33        & 94.65&\textbf{95.43 }      & 95.23 
  \\
                               & New      & 97.26  & 95.29   & {97.69} & 97.10      &  97.76&\textbf{97.76}       & 97.13  \\
                               & HM          & 94.12  & 94.47   & {96.43} & 95.69  &96.18    & \textbf{96.58 }       & 96.17 \\
\midrule
\multirow{3}{*}{\shortstack[l]{Stanford \\ Cars}}  & Base    & 63.37 & 78.12    & 70.49 &76.10 &71.76      & 72.94         & \textbf{78.80}  \\
                               & New     & 74.89 & 60.40    & 68.87 & 71.20        & \textbf{75.04}& 74.00      & 73.87   \\
                               & HM         & 68.65 & 68.13    & 72.01  &  72.30 &73.36       & 73.47         & \textbf{76.26} \\
\midrule
\multirow{3}{*}{Flowers102}    & Base     & 72.08 & 97.60    & 94.87  &97.23       &95.00 &95.92      &\textbf{96.83}   \\
                               & New     & \textbf{77.80} & 59.67    & 71.75 &69.27     &74.73 & 72.46        & 73.53   \\
                               & HM         & 74.83 & 74.06    & 81.71  & 80.90     &83.65& 82.56      & \textbf{83.59 } \\
\midrule
\multirow{3}{*}{Food101}       & Base     & 90.10 & 88.33    & {90.70}  & 90.37     &90.50 & \textbf{90.71 }      & 90.6 \\
                               & New     & 91.22 & 82.26    & 91.29  & 90.83       &91.70& \textbf{92.05}     & 91.33  \\
                               & HM          & 90.66 & 85.19    & 90.99  & 90.6    &91.09& \textbf{91.38}   &  90.96  \\
\midrule
\multirow{3}{*}{\shortstack[l]{FGVC \\ Aircraft}}  & Base    & 27.19 & 40.44    & 33.41 & 38.70     &36.21 & 37.44       & \textbf{42.47}  \\
                               & New      & \textbf{36.29} & 22.30    & 23.71 & 32.27      &33.55& {35.61}      &  33.47  \\
                               & HM         & 31.09 & 28.75    & 27.74  & 35.19    &34.83& 36.50   &  \textbf{37.43 } \\
\midrule
\multirow{3}{*}{SUN397}        & Base    & 69.36 & 80.60    &  79.74  &81.57        & 80.29 &80.82    &\textbf{81.97} \\
                               & New      & 75.35 & 65.89    &  76.86 &74.03    &76.53& \textbf{78.70 }        &  78.17 \\
                               & HM        & 72.23 & 72.51    &  78.27 &77.62    & 78.36&79.75    &  \textbf{80.02}  \\
\midrule
\multirow{3}{*}{DTD}           & Base    & 53.24 & 79.44    & 77.01 &79.53         &77.55 & 80.36      & \textbf{81.97}   \\
                               & New     & 59.90 & 41.18    & 56.00 &51.67      & 54.99&59.18       & \textbf{63.5} \\
                               & HM          & 56.37 & 54.24    & 64.85 &62.64   &64.35& 68.16      &  \textbf{71.56} \\
\midrule
\multirow{3}{*}{EuroSAT}       & Base    & 56.48 & 92.19    & 87.49  &87.70         &85.64 & \textbf{94.07}       & 93.33  \\
                               & New      & 64.05 & 54.74    & 60.04& 58.83      &64.34& \textbf{73.23  }    &  {71.07}   \\
                               & HM          & 60.03 & 68.69    & 71.21& 70.42 &73.48& \textbf{82.35}   & 80.69  \\
\midrule
\multirow{3}{*}{UCF101}        & Base     & 70.53  & 84.69   & 82.33  &85.47          &82.89 &83.00       &\textbf{85.23}   \\
                               & New     & 77.50  & 56.05   & 73.45  &72.97      &76.67 &\textbf{78.66}         & 77.3   \\
                               & HM          & 73.85  & 67.46   & 77.64 &78.73   &79.65& 80.77    & \textbf{81.07}  \\
\bottomrule
\end{tabular}%
}
    \caption{\small\textnormal{Accuracy comparison on Base-to-new generalization of CLIPFit with previous methods. Adapter: CLIP-Adapter.} }
    \label{tab:base-to-new}
    \vspace{-0.8em}
\end{table}
\textbf{Implementation details.}   We implemented our method with
PyTorch \cite{paszke2019pytorch}. The experiments were based on the vision 
backbone with Vit-B/16 \cite{dosovitskiy2020image}. We followed CoOp to preprocess input images. We used a single text prompt for all experiments for a fair comparison. We used SGD optimizer with batch size set as 32, and set the learning rate as 0.002 \cite{zhou2022learning}. All results
reported below are the average of three runs with different
random seeds. The training epoch was set to 100 for all datasets except ImageNet and Food101.  $\beta$ was set to 8 for all datasets on the base-to-new and cross-dataset setting, and 2 for the distribution shift setting. For the few-shot setting, we set $\beta$ to 2 for all datasets except SUN397 and DTD.  More implementation details are provided in appendix \ref{detail}.
 
\textbf{Comparisons.}
 We compared our method against state-of-the-art methods: zero-shot CLIP, prompt tuning methods: CoOp, CoCoOp \cite{zhou2022conditional}, ProGrad \cite{zhu2023prompt}, KgCoOp \cite{yao2023visual}, MaPLe \cite{khattak2023maple} and adapter tuning methods: CLIP-adapter, Tip-adapter \cite{zhang2021tip}. Detailed introductions to these methods can be found in appendix \ref{app:intro}.

\subsection{Comparisons with State-of-the-arts}\vspace{-0.4em}
\textbf{Results on base-to-new generalization setting}
Following \citet{zhou2022learning}, we split each dateset into two disjoint groups: the base class dataset and
the new class dataset. All compared methods and the proposed CLIPFit were trained on the base class dataset and evaluated on the new class dataset. We conducted $4/8/16$-shot experiments, following \citet{yao2023visual}. We reported base and
new class accuracies (Base and New) and their harmonic mean accuracy (HM). The 16-shot results are shown in Table \ref{tab:base-to-new}, and $4/8$-shot results are provided in appendix \ref{sec:detailed}. As shown in Table \ref{tab:base-to-new}, CLIPFit achieves 6 best HM accuracies among 11 datasets and the best average HM accuracy, which demonstrates that CLIPFit can not only learn well on seen base class data but also can generalize well to data from unseen new classes. A notable issue with previous methods like CoOp, CoCoOp,  ProGrad, and KgCoOp is that they usually perform well only on either the base or new class. To alleviate this issue, MaPLe proposes a multi-modal prompt
learning strategy for CLIP tuning which improves a lot over HM compared to previous methods.
Compared to MaPLe, our CLIPFit achieves better HM accuracy and average performance on the base class, with slightly lower average performance on the new class. It is important to note that CLIPFit only needs to tune nearly 46K parameters while MaPLe needs to tune nearly 3.55M parameters for each task, which is 77 times more than CLIPFit, meaning that CLIPFit fine-tunes significantly fewer parameters and is much more efficient.

\textbf{Results on few-shot learning setting.} 
 To verify whether the proposed CLIPFit can learn task-specific knowledge, we also compared CLIPFit with other existing methods on the few-shot learning setting. Following \citet{zhou2022conditional}, we used 1, 2, 4, 8, and 16-shot sets for training and reported accuracy performance. We report the results of the average accuracy of 11 datasets in Table. \ref{few-shot}, and report all results on each dataset in appendix \ref{more-few}. As shown in Table \ref{few-shot}, compared with other methods, CLIPFit shows overall consistent improvements among all $1/2/4/8/16$-shot settings. This demonstrates that CLIPFit can successfully learn task-specific knowledge. 
It is worth noting that CLIPFit outperforms other methods by a large margin in $1/2/4$-shot settings, demonstrating CLIPFit's robust ability to learn with extremely few
samples.

\textbf{Results on the robustness to distribution shift setting.}
Following \citet{zhang2021tip}, we evaluated the robustness under distribution shift of CLIPFit and other methods by first training models on the 16-shot ImageNet dataset and then evaluating on ImageNet-V2 \cite{recht2019imagenet} and  ImageNet-Sketch \cite{wang2019learning}. The label sets of two evaluating datasets are subsets of the label set of ImageNet. Although the label sets are compatible, the distributions of these three datasets are different from each other. The results are shown in Table \ref{tab:2label}. As shown in Table \ref{tab:2label}, while TIP-adapter achieves the best performance on the ImageNet dataset, CLIPFit can achieve better average performance compared to existing methods, effectively underlining the robustness of CLIPFit. 

Results on cross-dataset transfer
setting is provided in appendix \ref{cross-dataset}.

\begin{table}[t]
\renewcommand\arraystretch{0.9}
\footnotesize
\centering
\caption{Comparison of CLIPFit and other methods on the few-shot learning setting. We report average accuracy on 11 datasets for the $1/2/4/8/16$-shot setting.}
\resizebox{1\linewidth}{!}{
\begin{tabular}{c|ccccc}
\hline
\multirow{2}*{Method}&\multicolumn{5}{c}{shot}\\

&1&2&4&8&16\\
\hline
CoOP &68.09&70.13&73.59&76.45&79.01\\
CLIP-adapter &67.87&70.20&72.65&76.92&79.86\\
CoCoOp &66.95&67.63&71.98&72.92&75.02\\
ProGrad &68.2&71.78&74.21&77.93&79.2\\
KgCoOp &69.51&71.57&74.48&75.82&77.26\\
Tip-adapter &70.62&73.08&75.75&78.51&81.15\\ 
\hline
CLIPFit  &\textbf{72.32}&		\textbf{74.39}&		\textbf{77.18}	&	\textbf{79.03}	&	\textbf{81.27}\\

\hline
\end{tabular}}
\vspace{-0.8em}
\label{few-shot}
\end{table}

\begin{table}[t]
\renewcommand\arraystretch{0.9}
\footnotesize
\centering
\caption{Comparison of our method against other methods on robustness to distribution shift.}
\label{tab:2label}
\resizebox{1\linewidth}{!}{
\begin{tabular}{c|c|cc|c}
\hline
\multirow{2}*{Method}&Souce&\multicolumn{2}{c|}{Target}&\multirow{2}*{Average}\\
&ImageNet &-V2 &-Sketch \\
\hline
CLIP &66.73 &60.83 &46.15 &57.90\\
CoOP  &71.51& 64.20& 47.99& 61.23\\
CLIP-adapter & 71.60&63.67&46.52&60.60\\
Tip-adapter &\textbf{73.10}&{64.82}&46.73&61.55\\
CoCoOp & 71.02 &64.07 &48.75&61.28\\
ProGrad &{72.24} &64.73 &47.61&61.53\\
KgCoOp &71.20& 64.10& \textbf{48.97}& 61.42\\
\hline
CLIPFit  &71.53 &	\textbf{64.83}  &	48.87 &	\textbf{61.74}\\

\hline
\end{tabular}}
\vspace{-0.8em}
\end{table}

\subsection{Fine-tuning Analysis}
\label{sec:anan}
\begin{figure}[t]
\begin{center}
    \includegraphics[width=1\linewidth]{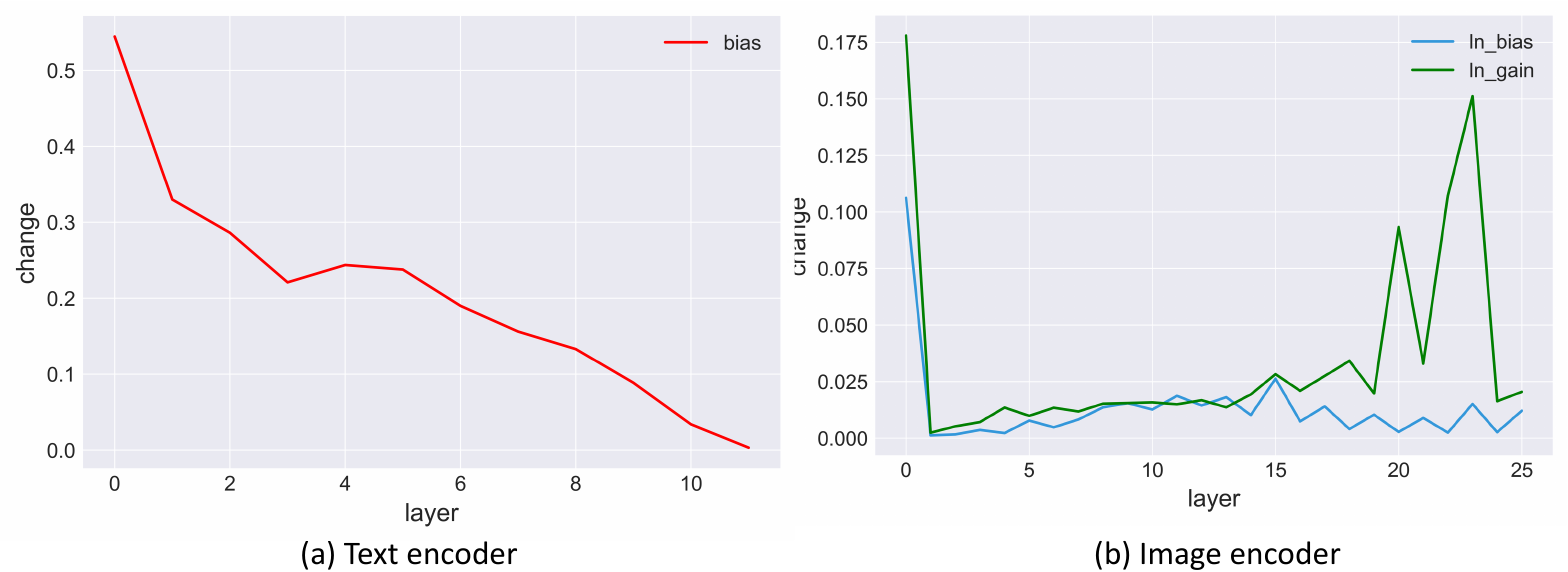}
\end{center}
\vspace{-0.1em}
\caption{Visualization of changes in different layers.
} 
\label{fig:change}
\vspace{-0.5em}
\end{figure}

\begin{figure}[t]
\begin{center}
    \includegraphics[width=1\linewidth]{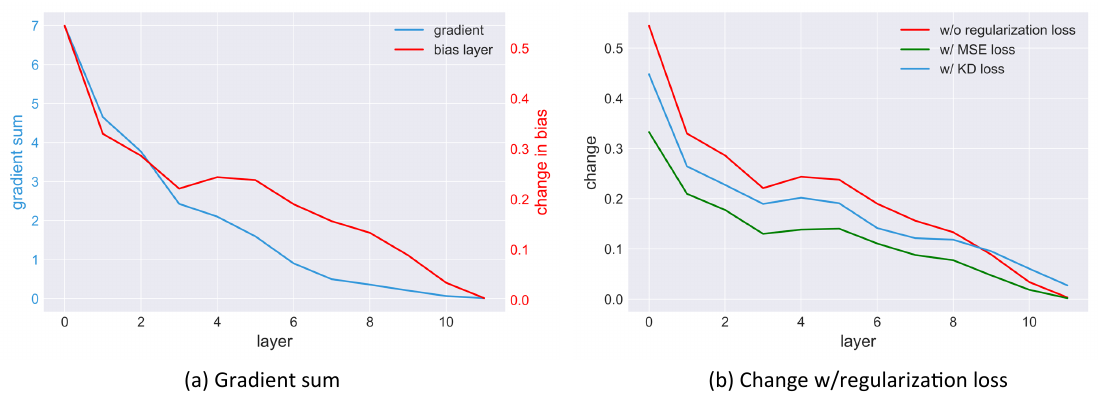}
\end{center}
\vspace{-0.1em}
\caption{Left: visualization of squared gradient sum. Right: visualization of change w/ regularization loss.
} 
\label{fig:change_loss}
\vspace{-0.5em}
\end{figure}

\textbf{Analyzing parameter change.} To understand the black-box fine-tuning process in CLIPFit, we first analyzed changes in the parameters of both the text encoder and image encoder. We computed the squared difference $||\boldsymbol{p}^\mathrm{pre}-\boldsymbol{p}||^2$ for each layer, where $\boldsymbol{p}^\mathrm{pre}$ is the pre-trained parameter vector and $\boldsymbol{p}$ is the fine-tuned parameter vector. We conduct experiments on the DTD dataset. The results are shown in Fig. \ref{fig:change}. As observed Fig. \ref{fig:change} (a), for bias terms in the FNN of the text encoder, when the number of layers increases, the change in bias decreases, which implies that low-level features in the text encoder change more than high-level features during the fine-tuning process of CLIPFit. From Fig. \ref{fig:change} (b), we found that for LayerNorm in the image encoder, the first layer (i.e., patch embedding layer) changes much more compared with other layers for both bias and gain, 
showing that tuning patch embedding LayerNorm is crucial for shifted downstream tasks. Moreover, the gain of the last several LayerNorm layers has much changed and the most intermediate layers change much less. The difference in change between different layers may be caused by gradient difference. We visualize the squared sum of gradient from each text bias layer in Fig. \ref{fig:change_loss} (a). As observed, the curve of the gradient sum is very similar to changes in parameters. 

To verify whether more changed layers are more important in fine-tuning, we conducted experiments by freezing less (or more) changed LayerNorm layers on the 4-shot setting. We found that when only updating the first LayerNorm layer and freezing other LayerNorm layers, the average accuracy is 76.22\%. For comparison, the average accuracy is 74.93\% when only updating the twelfth LayerNorm, and 75.06\% when updating the last LayerNorm. Both are much less than the first layer and these two layers change much less than the first layer, as shown in Fig. \ref{fig:change} (b). Moreover, when updating the top 6 most changed LayerNorm layers, the average accuracy is 77.03\%, which is only a 0.15\% drop, while only tuning 23\% parameters of CLIPFit. The phenomenon for text encoder is similar and can be found in appendix \ref{more anan}.  These results demonstrate that the more changed layers are more important for knowledge adapting.

\textbf{Analyzing regularization loss.} We then analyze the two regularization losses: KD loss and bias term MSE loss. We found that both two losses can avoid overfitting and boost performance during fine-tuning. For the 4-shot learning task, fine-tuning w/ KD loss leads to a 77.18\% average accuracy, and fine-tuning w/ KD loss leads to a 76.23\% average accuracy. Both two performances are better than fine-tuning w/o regularization loss (76.13\% average accuracy) and KD loss performs better. We then analyze how these two losses affect changes in parameters during fine-tuning of CLIPFit. The results are shown in Fig. \ref{fig:change_loss} (b). As observed, KD loss will reduce the changes for the more-changed low-level bias terms and enhance changes in less-changed high-level layers, which implies that penalizing changes for low-level bias terms is important in avoiding overfitting. Compared with KD loss, MSE loss directly applying to text bias terms reduces more changes in low-level layers. 

\begin{figure}[t]
\begin{center}
    \includegraphics[width=1\linewidth]{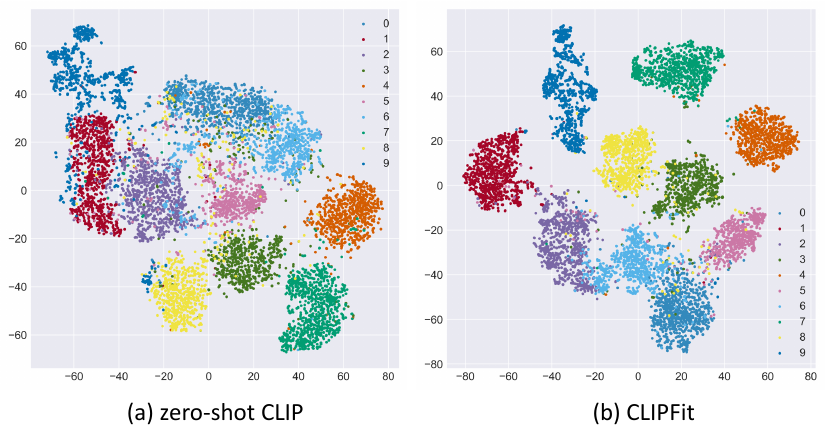}
\end{center}
\vspace{-0.1em}
\caption{Visualization of learned image feature space from zero-shot CLIP and CLIPFit via t-SNE.
} 
\label{fig:t-sne}
\vspace{-0.7em}
\end{figure}

\textbf{Analysing image encoder representations.}
We used t-SNE \cite{van2008visualizing} to visualize the image representation space of zero-shot CLIP and CLIPFit to analyze image encoder representations. We visualize the data from EuroSAT dataset. The visualization results are presented in Fig. \ref{fig:t-sne}. As observed, in high-dimensional classification feature space, CLIPFit has a much clearer separation of different class image features compared with zero-shot CLIP, which demonstrates that CLIPFit can better detect the similarities among images. These results verify that updating LayerNorm in the image encoder during fine-tuning will lead to a more separated and better similarity-detected image feature space.

\begin{table}[t]
\renewcommand\arraystretch{0.9}
\footnotesize
\centering
\caption{Comparison of prompt tuning and adapter tuning methods w/ and w/o updating LayerNorm on the 16-shot base-to-new setting. +UL means training  \underline{u}pdating \underline{L}ayerNorm.}
\label{tab:adding}
\resizebox{1\linewidth}{!}{

\begin{tabular}{c|cc|c}
\hline
Method&Base&New&H\\
\hline
CoOp&82.63 &67.99 &74.60\\
+UL&82.96(\textcolor{red}{+0.33})	&69.09(\textcolor{red}{+1.10})	&75.39(\textcolor{red}{+0.79})\\
\hline
KgCoOp &80.73 &73.60 &77.00\\
+UL &82.13(\textcolor{red}{+1.40})	&74.96(\textcolor{red}{+1.36})  &78.38(\textcolor{red}{+1.38})\\
\hline
CLIP-adapter& 82.23& 70.61& 75.98\\
+UL &83.63(\textcolor{red}{+1.40})	&71.87(\textcolor{red}{+1.26})	&77.31(\textcolor{red}{+1.33})\\
\hline
\end{tabular}
}
\vspace{-0.8em}
\end{table}

\textbf{Updating LayerNorm can also benefit other methods.} We show that updating LayerNorm can also benefit prompt tuning methods and adapter tuning methods. We re-implemented CoOp,
KgCoOp, and CLIP-adapter with updating LayerNorm. The results are shown in Table \ref{tab:adding}. Table \ref{tab:adding} shows that training with LayerNorm updating can boost the base class, new class, and harmonic mean performance for all three methods. For example, training KgCoOp
 with updating LayerNorm can bring 1.4\%, 1.36\%, and 1.38\% improvements in the base class, new class, and Harmonic Mean (HM) accuracy, which demonstrates the effectiveness and wide validity of the proposed updating LayerNorm.

 More detailed analyses about other datasets and other aspects are provided in appendix \ref{more anan}.
 \begin{table}[t]
\renewcommand\arraystretch{0.9}
\footnotesize
\centering
\caption{Comparison of different strategies of fine-tuning bias terms in CLIP.}
\label{tab:biasana}
\resizebox{0.95\linewidth}{!}{
\begin{tabular}{c|ccc|c}
\hline
Strategy&Base&New&H&\# param.\\
\hline
(a) Text+Image bias&84.15&	64.35&	72.93&0.17M\\
(b) Text bias&83.33&	64.43&	72.67&67.6K\\
(c) FFNs bias (Text)&83.25&	67.60&	74.61&30.7K \\
(d) Projection bias (Text)&83.23& 67.58& 74.59& 6.1K\\
\hline
\end{tabular}
}
\vspace{-1em}
\end{table}
\subsection{Ablation Study}
\label{sec:abla}
\textbf{Comparison of different strategies of fine-tuning bias terms.}
We give an in-depth exploration of how to apply BitFit to fine-tune the CLIP model. Original BitFit fine-tunes all bias terms in language models. We conduct 4 strategies for fine-tuning bias terms of CLIP: (a) fine-tuning all bias terms of the text and image encoder; (b) fine-tuning all bias terms of the text encoder; (c) fine-tuning bias terms of FFNs of the text encoder; (d) fine-tuning bias terms of projection linear layers in FFNs of the text encoder. We trained these four strategies on the 16-shot base-to-new setting with $\mathcal{L}_\mathrm{ce}$ and reported average accuracy. The results are shown in Table \ref{tab:biasana}. As shown in Table \ref{tab:biasana},  both strategy (a) and strategy (b) can boost seen base class performance but will decrease significantly unseen new class performance, which implies that directly applying BitFit to CLIP may be harmful to model's generalization ability. Moreover, strategy (c)  and strategy (d) can have similar performance among both the base and new class data, but strategy (d) fine-tunes only one-fifth of parameters compared with strategy (c), which speeds up training.

\begin{table}[t]
\vspace{0pt}
\renewcommand\arraystretch{0.9}
\footnotesize
\centering
\resizebox{0.95\linewidth}{!}{
\begin{tabular}{cccc}
\hline
\multirow{2}*{Configurations}&\multicolumn{3}{c}{Accuracy(\%)}\\

&1-shot&4-shot&16-shot\\
Porjection Bias&68.86&74.72	&79.53\\
w/ LayerNorm&70.75&	76.13&	81.04\\
w/ KD loss&71.21	&75.94	&79.62\\
w/ KD loss + LayerNorm&72.32 &77.18 &81.27\\
\hline
\end{tabular}
}
\captionof{table}{Ablation study of CLIPFit on the few-shot setting with $1/4/16$-shot. Projection Bias: fine-tuning bias terms of projection layer in FNNs of the text encoder. LayerNorm: updating LayerNorm in the image encoder.}
\label{tab:ablation}
\end{table}
\textbf{Effectiveness of proposed components.}
We validated the
effects of updating LayerNorm and KD loss by ablating them. The results are shown in Table \ref{tab:ablation}. Fine-tuning bias terms with KD loss brings 2.35\%, 1.22\%, and 0.09\% improvements for $1/4/16$-shot setting, respectively. Fine-tuning bias terms in the text encoder and LayerNorm in the image encoder brings 1.89\%, 1.41\%, and 1.51\% improvements for $1/4/16$-shot setting, respectively. Together, CLIPFit brings 3.46\%, 2.46\% and 1.74\% improvements for $1/4/16$-shot setting, respectively. These results demonstrate the effectiveness of each CLIPFit components.

\begin{table}[t]
\vspace{0pt}
\renewcommand\arraystretch{0.9}
\footnotesize
\centering
\resizebox{0.95\linewidth}{!}{
\begin{tabular}{ccc|ccc}
\hline
Method&Params&time&Base&New&HM\\
CoOp&2048&0.44ms &82.69&63.22&71.66\\
CoCoOp&35k&25.59ms & 80.47&71.69 &75.83\\
KgCoOp &2048&0.44ms&80.73&73.60&77.00\\
MaPLe &3.55M&2.1ds&82.28&75.14&78.55\\
CLIPFit &44k&0.96ms&83.72&74.84&79.03\\
\hline
\end{tabular}
}
\captionof{table}{Comparison of training efficiency with other
methods over 11 datasets.}
\vspace{-0.3em}
\label{tab:effciency}
\end{table}
\textbf{Training efficiency.} We compare the training efficiency of CLIPFit and other methods w.r.t.  parameters and training time per image \cite{yao2023visual}. The results are shown in Table \ref{tab:effciency}.  It is noticed that CoOp and KgCoOp have the lowest number of training parameters and time. However, the performance of these two methods is not satisfactory. MaPLe improves accuracy performance compared with other methods but also increases the required tuning parameters to 3.55M, which is very time-consuming.  CLIPFit achieves the best harmonic mean accuracy with only 44k parameters, which is much less than MaPLe. Also, the training time of CLIPFit is slightly higher than CoOp and KgCoOp. Given the large improvement of CLIPFit, a slight increase in training time is acceptable.
\subsection{Discussion}
Although our method is designed for contrastive encoder VLMs (CLIP), the core idea of CLIPFit and our model analysis may still provide insights for other large multimodal model (e.g., LLaVA \cite{liu2024visual}) fine-tuning and many furthur applications \cite{touvron2023llamaopenefficientfoundation, mizrahi20234mmassivelymultimodalmasked, geminiteam2024geminifamilyhighlycapable, geminiteam2024gemini15unlockingmultimodal, li2024dpudynamicprototypeupdating, 10447193, 10.1145/3581783.3611847, zhang2022learningfreeobjectsegments, chen2023federatedlearningshareablebases, Li_2023_CVPR}. For example, the idea of tuning LayerNorm could be used when distributions of downstream and pretraining QA image data are divergent, and parameter change and importance analysis (Sec. \ref{sec:anan}) could provide insights for how to select fine-tuning parameters. We hope our work can provide insights for a broader range of VLM fine-tuning.
\section{Conclusion}
In this paper, we presented CLIPFit for fine-tuning visual-language models. Unlike existing prompt tuning and adapter tuning methods, CLIPFit does not introduce any external parameters and fine-tunes CLIP by updating only bias terms of projection layers in FFNs of the text encoder and the image encoder's LayerNorm. To understand the effect of CLIPFit fine-tuning on the pre-trained model, we conducted various analyses focusing on changes in internal parameters and representations. We conducted extensive experiments and analysis to evaluate CLIPFit on 11 datasets, whose performances show the superiority of our method.

\section{Limitations}
In this paper, we presented CLIPFit for VLM fine-tuning and conducted an exploration of how CLIPFit affects the pre-trained CLIP model. Our analyses found some interesting phenomena after fine-tuning, i.e., low-level bias terms in the text encoder change much more than high-level bias terms and the change in the first LayerNorm layer is much bigger than other LayerNorm layers in the image encoders. Moreover, we found that this may be caused by the difference in the magnitude of the gradient. Nevertheless, our analysis does not reveal why the difference in the magnitude of the gradient happens during fine-tuning. A deeper analysis of gradient back-propagation during fine-tuning is needed to understand this for future work.

Furthermore, following previous works \cite{zhou2022conditional,yao2023visual,zhou2022learning,khattak2023maple}, this paper focused on image classification for VLMs, so our study was constrained to classification tasks. Expanding CLIPFit for VLM fine-tuning to a broader range of tasks (e.g., image retrieval) could be the future work. 

\section*{Acknowledgement}
MS was partially supported by Institute for AI and Beyond, UTokyo.
\bibliography{main}

\appendix
\newpage

\section{Dataset Statistics}
\label{sec:data}
Following CoOp \cite{zhou2022learning}, we conducted extensive experiments on 11 public classification benchmark datasets to evaluate the effectiveness of the proposed CLIPFit. The datasets are  ImageNet \cite{deng2009imagenet}, Caltech101 \cite{fei2004learning},
OxfordPets \cite{parkhi2012cats}, StanfordCars \cite{krause20133d}, Flowers102 \cite{nilsback2008automated},
Food101 \cite{bossard2014food}, FGVCAircraft \cite{maji2013fine}, SUN397 \cite{xiao2010sun}, DTD \cite{cimpoi2014describing},
EuroSAT \cite{helber2019eurosat}, and UCF101 \cite{soomro2012ucf101}. In distribution shift experiments, we also used ImageNet-V2 \cite{recht2019imagenet}, and  ImageNet-Sketch \cite{wang2019learning} as the target dataset. The statistics of these datasets can be found in Table \ref{tab:datasets}.

\section{Implementation Details}
\label{detail}
We implemented our method with
PyTorch \cite{paszke2019pytorch}. All experiments were based on the vision 
backbone with Vit-B/16 \cite{dosovitskiy2020image} of CLIP \cite{radford2021learning}. We followed CoOp \cite{zhou2022learning} to preprocess input images: we resized all images to 224×224 and used random
cropping, resizing, and random horizontal flipping for data augmentation.
Following \citet{radford2021learning}, we used a single hand-craft prompt as text input for all methods except prompt tuning methods for a fair comparison. The prompt for each dataset can be found in Table \ref{tab:datasets}.  We used SGD optimizer with batch size set as 32, and set the learning rate as 0.002 \cite{zhou2022learning}. All results
reported below are the average of three runs with different
random seeds. The training epoch was set to 100 for all datasets except ImageNet and Food101. The training epoch for ImageNet and Food101 datasets was set to 10. Smoothing parameter $\alpha$ was set to 0.99 for all experiments. $\beta$ was set to 8 for all datasets on the base-to-new and cross-dataset setting, and 2 for the distribution shift setting. For the few-shot setting, we set $\beta$ to 2 for all datasets except SUN397 and DTD. $\beta$ was set to 8 for SUN397 and DTD datasets. All experiments were run on one single NVIDIA A100 GPU.
\begin{figure}[t]
\centering
\begin{center}
    \includegraphics[width=0.6\linewidth]{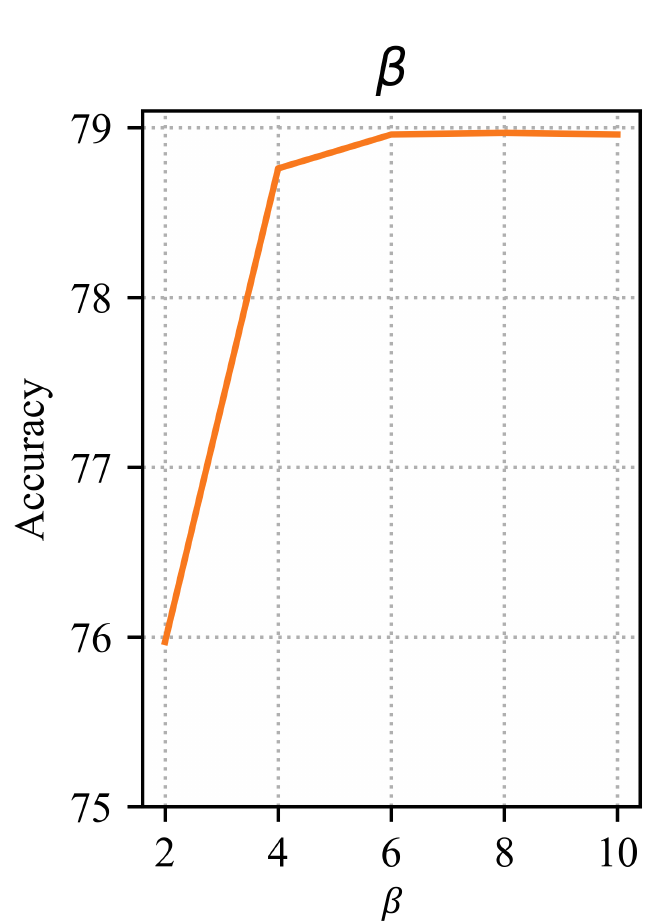}
    \end{center}
\caption{Performance changes of harmonic mean on 16 shot base-to-new setting by varying
hyperparameter $\beta$.}.
\label{fig:para}
\end{figure}
\section{Detailed Introduction to Baseline Methods.}
\label{app:intro}
We compared our method against state-of-the-art methods: zero-shot CLIP \cite{radford2021learning}, prompt tuning methods: CoOp \cite{zhou2022learning}, CoCoOp \cite{zhou2022conditional}, ProGrad \cite{zhu2023prompt}, KgCoOp \cite{yao2023visual} , MaPLe \cite{khattak2023maple} and adapter tuning methods: CLIP-ddapter \cite{gao2023clip}, Tip-adapter \cite{zhang2021tip}. 

Zero-shot CLIP \cite{radford2021learning} uses the hand-crafted template “a photo
of a []” to generate the prompts and then applies these prompts to predict the class of given images.

CoOp \cite{zhou2022learning} introduces learnable text prompts instead of hand-crafted prompts to adapt the CLIP model to downstream image recognition tasks. 

CoCoOp \cite{zhou2022conditional} proposes to generate an input-conditional token for each image with a lightweight learnable neural network.

KgCoOp \cite{yao2023visual} proposes to 
use context knowledge distillation to learn from the original CLIP model to avoid overfitting and forgetting.

 MaPLe \cite{khattak2023maple} proposes a multi-modal prompt learning strategy to introduce learnable text and image prompts.
 
CLIP-Adapter \cite{gao2023clip} sets an additional bottleneck layer following the text or image encoder to learn better features by a residual way.

Tip-Adapter\cite{zhang2021tip} does not need training but creates the weights by a key-value cache model constructed from the few-shot training set and then uses this cache model for inference.
\begin{table*}[t]
    \centering
    \caption{Statistics and prompts for each Dataset.}
    \label{tab:datasets}
    \begin{tabular}{l r r r r c}
    \toprule
    Dataset & Classes & Train & Val & Test & Hand-crafted prompt \\
    \midrule
    ImageNet & 1,000 & 1.28M & N/A & 50,000 & ``a photo of a [CLASS].'' \\
    Caltech101 & 100 & 4,128 & 1,649 & 2,465 & ``a photo of a [CLASS].'' \\ 
    OxfordPets & 37 & 2,944 & 736 & 3,669 & ``a photo of a [CLASS], a type of pet.'' \\
    StanfordCars & 196 & 6,509 & 1,635 & 8,041 & ``a photo of a [CLASS].'' \\
    Flowers & 102 & 4,093 & 1,633 & 2,463 & ``a photo of a [CLASS], a type of flower.'' \\
    Food101 & 101 & 50,500 & 20,200 & 30,300 & ``a photo of [CLASS], a type of food.'' \\
    FGVCAircraft & 100 & 3,334 & 3,333 & 3,333 & ``a photo of a [CLASS], a type of aircraft.'' \\
    SUN397 & 397 & 15,880 & 3,970 & 19,850 & ``a photo of a [CLASS].'' \\
    DTD & 47 & 2,820 & 1,128 & 1,692 & ``[CLASS] texture.'' \\
    EuroSAT & 10 & 13,500 & 5,400 & 8,100 & ``a centered satellite photo of [CLASS].'' \\
    UCF101 & 101 & 7,639 & 1,898 & 3,783 & ``a photo of a person doing [CLASS].'' \\
    \midrule
    ImageNetV2 & 1,000 & N/A & N/A & 10,000 & ``a photo of a [CLASS].'' \\
    ImageNet-Sketch & 1,000 & N/A & N/A & 50,889 & ``a photo of a [CLASS].'' \\
    \bottomrule
    \end{tabular}
\end{table*}

\begin{table*}[t]
    \tabstyle{5pt}
    \caption{Comparison of CLIPFit and other methods in the cross-dataset transfer setting. S.C.: StanfordCars dataset. F.A.: FGVCAircraft dataset.
    }
    \label{tab:cd}
    \begin{tabular}{l c ccccccccccc}
    \toprule
    & Source & \multicolumn{11}{c}{Target} \\ \cmidrule(lr){2-2} \cmidrule(lr){3-13}
    & \rotatebox{55}{ImageNet} & \rotatebox{55}{Caltech101} & \rotatebox{55}{OxfordPets} & \rotatebox{55}{S.C.} & \rotatebox{55}{Flowers102} & \rotatebox{55}{Food101} & \rotatebox{55}{F.A.} & \rotatebox{55}{SUN397} & \rotatebox{55}{DTD} & \rotatebox{55}{EuroSAT} & \rotatebox{55}{UCF101} & \rotatebox{55}{\emph{Average}} \\
    \midrule
    CoOp  & {71.51} & 93.70 & 89.14 & 64.51 & 68.71 & 85.30 & 18.47 & 64.15 & 41.92 & {46.39} & 66.55 & 63.88 \\
    CoCoOp & 71.02 & \textbf{94.43} &{90.14} & {65.32} & \textbf{71.88} & {86.06} & {22.94} & {67.36} & {45.73} & {45.37} & {68.21} & {65.74} \\
    ProGrad & \textbf{72.24}&91.52&89.64&62.39&67.87&85.40&20.61&62.47&39.42&43.46&64.29&62.71\\
    KgCoOp&70.66&93.92&89.83&\textbf{65.41}&70.01&\textbf{86.36}&22.51&66.16&\textbf{46.35}&46.04&68.50&65.51\\
\midrule
    CLIPFit&71.10&93.77&\textbf{90.36}&64.56 &71.43&85.76&\textbf{24.46} &\textbf{67.43 }&45.20&\textbf{46.40}&\textbf{69.17}&\textbf{65.85}\\

    \bottomrule
    \end{tabular}
    \vspace{-0.8em}
\end{table*}

\section{Results on cross-dataset transfer setting}
\label{cross-dataset}
Following \citet{zhou2022learning,zhou2022conditional}, We also evaluated the cross-dataset generalization ability of CLIPFit and other methods. Models were trained on the 16-shot Imagenet dataset and then tested on other datasets. The results are shown in Table \ref{tab:cd}. As shown in Table \ref{tab:cd}, the average performance of CLIPFit is also better than existing methods, which shows that CLIPFit has a good generalization ability.   \vspace{-0.2em}

\section{Parameter Analysis}
\label{alation}
 In this section, we aim to discuss hyper-parameters $\beta$. $\alpha$ is the coefficient parameter to control the weight of knowledge distillation loss. Experiments were conducted on the 16-shot base-to-new setting, and we report harmonic mean accuracy in Fig. \ref{fig:para}. As shown in Fig.~\ref{fig:para}, performances are not sensitive within certain ranges.

\section{More Few-shot Learning Results}
\label{more-few}
Following \citet{zhou2022conditional}, we used 1, 2, 4, 8, and 16-shot sets for training and reported accuracy performance to test whether our proposed method can learn task-specific knowledge. The results are reported in Table \ref{tab:fewshot}. As shown in Table \ref{tab:fewshot}, CLIPFit can bring a consistent improvement in terms of average accuracy on all settings.
\label{sec:appendix}

\section{More Results of base-to-new setting}
In this section, we give more detailed results on the base-to-new setting. The detailed results for each dataset
on 4-shot and 8-shot settings are shown in Table \ref{tab:vit-4} and Table \ref{tab:vit-8}. Since  MaPLe \cite{khattak2023maple} did not conduct experiments on $4/8$-shot setting, we do not report results from MaPLe. As shown in Table \ref{tab:vit-4} and Table \ref{tab:vit-8}, the proposed CLIPFit brings consistent improvement compared with other methods. 
\label{sec:detailed}

\begin{figure}[t]
\begin{center}
    \includegraphics[width=1\linewidth]{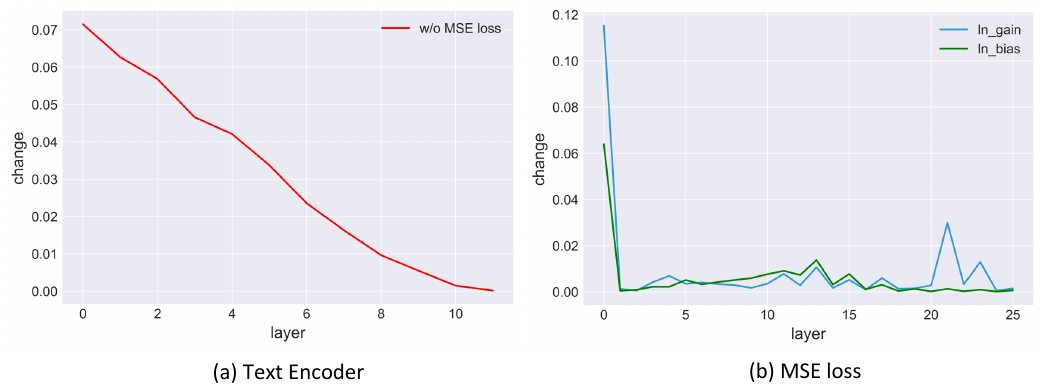}
\end{center}
\vspace{-0.1em}
\caption{Visualization of changes in different layers on the EuroSAT dataset.
} 
\label{fig:change_1}
\vspace{-0.5em}
\end{figure}

\begin{figure}[t]
\begin{center}
    \includegraphics[width=1\linewidth]{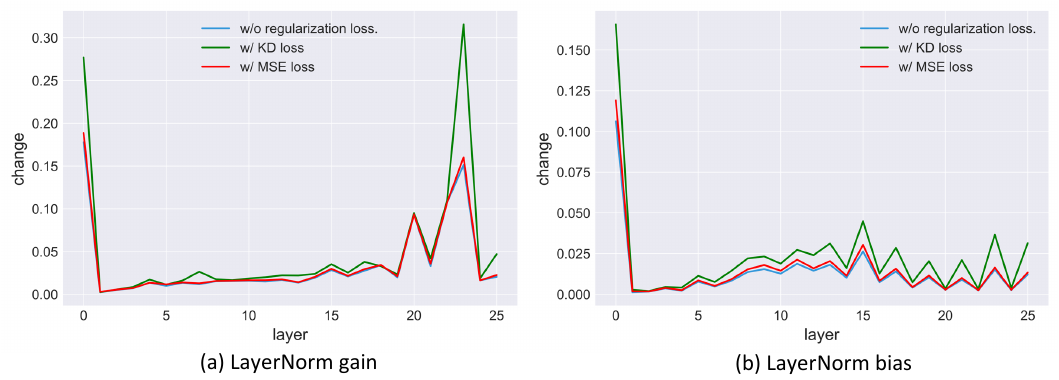}
\end{center}
\vspace{-0.1em}
\caption{Visualization of LayerNorm changes w/ and w/o regularization loss in the DTD dataset.
} 
\label{fig:change_loss_layernorm_dtd}
\vspace{-0.5em}
\end{figure}

\begin{figure}[t]
\begin{center}
    \includegraphics[width=1\linewidth]{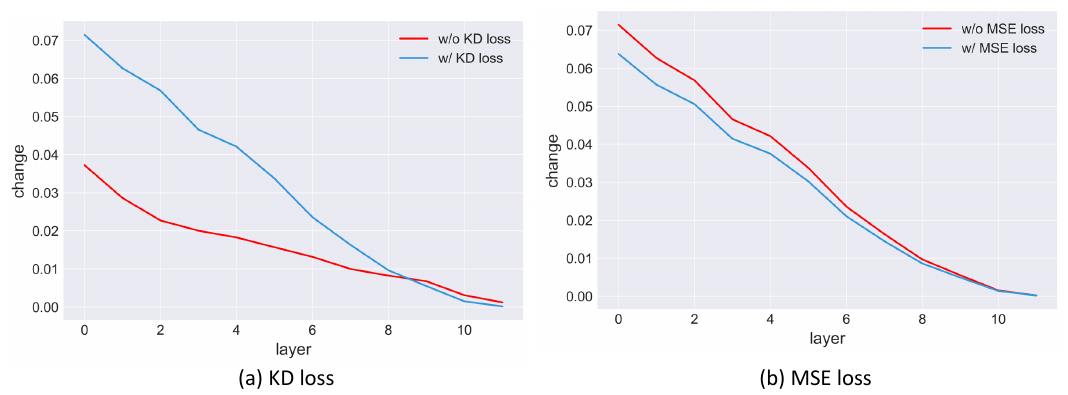}
\end{center}
\vspace{-0.1em}
\caption{Visualization of bias changes w/ and w/o regularization loss in the EuroSAT dataset.
} 
\label{fig:change_loss_2}
\vspace{-0.5em}
\end{figure}

\begin{figure}[t]
\begin{center}
    \includegraphics[width=1\linewidth]{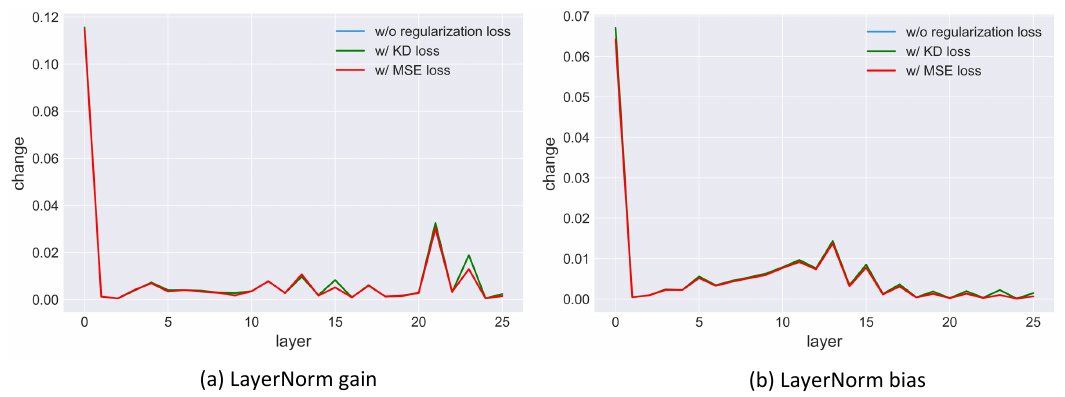}
\end{center}
\vspace{-0.1em}
\caption{Visualization of LayerNorm changes w/ and w/o regularization loss in the EuroSAT dataset.
} 
\label{fig:change_loss_3}
\vspace{-0.5em}
\end{figure}

\begin{figure}[t!]
\begin{center}
    \includegraphics[width=1\linewidth]{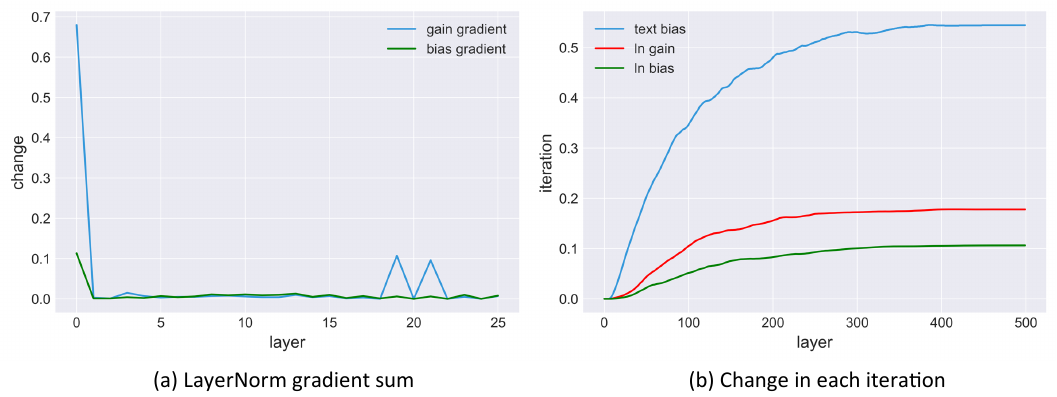}
\end{center}
\vspace{-0.1em}
\caption{Left: visualization of squared gradient sum in LayerNorm layers. Right: change of the first text bias layer and first LayerNorm layer at each iteration.
} 
\label{fig:change_iteration}
\vspace{-0.5em}
\end{figure}

\section{More Fine-tuning Analysis}
\label{more anan}
Sec. \ref{sec:anan} discussed the changes in unfixed parameters after fine-tuning the DTD dataset and the importance of more changed LayerNorm. In this subsection, we give more detailed analyses of other datasets and other aspects.

\textbf{Importance of low-level bias terms in text encoder.} Sec. \ref{sec:anan} presented that after the fine-tuning of CLIPFit, for bias terms in the FNN of the text encoder, as the number of layers increases, the change in bias decreases. In this subsection, we conducted experiments to verify whether more changed layers in the text encoder are more important. Similar to Sec. \ref{sec:anan}, we freeze less (or more) changed LayerNorm bias layers in the text encoder on the 4-shot setting. When updating only the first bias layer and freezing other layers, the average accuracy is 74.69\%. For comparison, the average accuracy is 73.33\% when only updating the sixth bias layer and 70.62\% when only updating the last bias layer. Both are much lower than updating the first layer. We also found that when only updating the top-3 bias layers (changed more) and freezing other bias term layers, the average accuracy is 76.13\%. For comparison, when only updating the last 3 bias layers (changed less) and freezing other bias term layers, the average accuracy is 70.86\%, which is much lower than updating the top 3 bias layers.
These results demonstrate that the more changed parameters are crucial for fine-tuning.

\textbf{Analyzing LayerNorm with regularization loss.} Sec. \ref{sec:anan} analyzed the difference of changes in the text encoder bias terms between w/ and w.o regularization loss. In this subsection, we will analyze LayerNorm in the image encoder bias terms between w/ and w.o regularization loss. Noted that although the two regularization losses are applied to text features or text encoder, the image encoder or image features will also be affected since these two encoders are fine-tuned simultaneously. The results on the DTD dataset are shown in Fig. \ref{fig:change_loss_layernorm_dtd}. When fine-tuning w/ KD loss, unlike in text encoder, changes in gain and bias increase compared with w/o KD loss. This phenomenon implies that image features will change more w/ KD loss compared with fine-tuning w/o KD loss. Moreover, we also found that the increases are almost in the more changed LayerNorm layers. When fine-tuning w/ MSE loss, changes in gain and bias are equal or slightly higher than fine-tuning w/o KD loss.

\begin{table*}[t]
\caption{Comparison with existing methods in the few-shot learning setting. S.C.: StandfordCars dataset. F.A.: FGVCAircraft dataset.}
\label{tab:fewshot}
\centering
\small
\resizebox{1.\linewidth}{!}{
\begin{tabular}{c|c|ccccccccccc|c}
\toprule
              shot&Method&ImageNet& {Caltech101}  &OxfordPets&S.C.&Flowers&Food101&F.A.&Sun397&DTD&EuroSAT&UCF101&AVG                                            \\
              \midrule
              \multirow{8}{*}{1}& CoOp&65.77&92.37&\textbf{92.2}&67.1&82.2&82.07&26.73&64.7&49.0&54.8&72.0&68.09\\
&CoCoOp&\textbf{69.51}&{93.8}&91.17&67.92&71.98&86.1&13.2&68.19&48.51&55.71&70.35&66.95\\
&CLIP-adapter&67.93&93.3&89.03&67.1&72.03&85.9&27.6&67.1&45.2&61.7&69.67&67.87\\
&TIP-adapter&67.43&93.56&90.72&67.88&\textbf{86.63}&86.01&\textbf{29.58}&64.49&53.25&63.95&73.27&70.62\\
&ProGrad&64.33&90.96&89.01&67.11&{83.81}&82.75&{27.97}&64.54&52.79&55.1&71.91&68.2\\
&KgCoOp&69.03&\textbf{94.13}&{91.97}&67.03&74.63&\textbf{86.27}&26.9&68.43&52.5&60.83&72.93&69.51\\
&CLIPFit&69.37&93.67&91.63&\textbf{69.33}&82.83&86.17&27.73&\textbf{69.07}&\textbf{54.63}&\textbf{76.87}&\textbf{74.27}&\textbf{72.32}\\

\bottomrule   
\bottomrule
\multirow{8}{*}{2}& CoOp&68.17&92.83&89.2&69.37&88.47&80.8&29.57&66.4&51.7&61.2&73.67&70.13\\
&CoCoOp&69.84&\textbf{94.92}&\textbf{92.14}&68.77&76.12&86.21&15.03&69.11&52.02&46.24&73.58&67.63\\
&CLIP-adapter&68.6&93.67&89.73&68.97&78.53&86.1&29.6&69.0&47.87&66.07&74.1&70.2\\
&TIP-adapter&68.6&94.22&91.1&71.39&\textbf{90.21}&86.26&\textbf{32.51}&66.74&56.32&{70.38}&76.11&73.08\\
&ProGrad&66.12&93.21&90.55&71.94&{88.62}&84.81&{30.84}&68.51&54.35&66.19&74.39&71.78\\
&KgCoOp&69.63&94.2&92.13&68.13&79.47&{86.6}&28.07&69.53&55.73&68.97&74.83&71.57\\
&CLIPFit&\textbf{69.93}&94.47&92.03&\textbf{72.7}&87.77&\textbf{86.63}&30.7&\textbf{70.87}&\textbf{57.7}&\textbf{78.83}&\textbf{76.7}&\textbf{74.39}
\\
\bottomrule   
\bottomrule
\multirow{8}{*}{4}& CoOp&69.38&94.44&91.3&72.73&91.14&82.58&33.18&70.13&58.57&68.62&77.41&73.59\\
&CoCoOp&\textbf{70.55}&94.98&93.01&69.1&82.56&\textbf{86.64}&30.87&70.5&54.79&63.83&74.99&71.98\\
&CLIP-adapter&69.56&94.0&90.87&71.13&86.77&86.47&31.1&71.3&53.83&66.8&77.3&72.65\\
&TIP-adapter&69.86&\textbf{95.06}&91.58&74.59&91.5&86.48&{35.15}&70.29&62.09&{76.43}&\textbf{80.24}&75.75\\
&ProGrad&70.21&94.93&\textbf{93.21}&71.75&89.98&85.77&32.93&71.17&57.72&70.84&77.82&74.21\\
&KgCoOp&70.19&94.65&{93.2}&71.98&90.69&86.59&32.47&71.79&58.31&71.06&78.4&74.48\\
&CLIPFit&70.4&95.0&93.07&\textbf{76.43}&\textbf{92.03}&86.73&\textbf{35.8}&\textbf{73.0}&\textbf{63.2}&\textbf{83.17}&{80.13}&\textbf{77.18}\\

\bottomrule   
\bottomrule
\multirow{8}{*}{8}& CoOp&70.83&94.1&90.83&76.57&94.37&83.37&37.63&72.5&64.7&75.53&80.57&76.45\\
&CoCoOp&70.77&95.11&\textbf{93.44}&70.19&84.17&86.92&26.53&70.62&58.92&68.26&77.19&72.92\\
&CLIP-adapter&70.2&94.27&91.77&76.47&94.9&86.67&37.2&73.13&66.4&73.23&81.87&76.92\\
&TIP-adapter&\textbf{71.4}&95.2&92.09&78.34&\textbf{94.98}&86.74&\textbf{40.61}&73.6&\textbf{67.28}&{81.11}&{82.24}&{78.51}\\
&ProGrad&{71.1}&94.92&92.18&{78.78}&93.51&85.91&37.89&72.91&62.13&79.22&\textbf{88.64}&77.93\\
&KgCoOp&70.23&94.97&93.1&73.53&89.53&86.9&34.97&72.5&65.87&72.37&80.03&75.82\\
&CLIPFit&71.0&\textbf{95.43}&93.13&\textbf{79.57}&94.7&\textbf{87.0}&39.93&\textbf{74.27}&67.17&\textbf{84.87}&82.17&\textbf{79.03}\\
\bottomrule   
\bottomrule
\multirow{8}{*}{16}& CoOp&71.51&95.5&91.8&78.89&96.1&85.17&40.93&74.5&68.63&83.6&82.43&79.01\\
&CoCoOp&71.02&95.19&93.25&71.68&87.64&87.19&31.29&72.05&63.78&73.82&78.34&75.02\\
&CLIP-adapter&71.6&94.57&92.03&80.9&\textbf{97.0}&86.83&42.67&75.3&71.17&81.87&\textbf{84.53}&79.86\\
&TIP-adapter&\textbf{73.1}&95.79&92.7&\textbf{83.09}&96.18&\textbf{87.24}&\textbf{45.59}&74.99&\textbf{72.05}&{87.46}&{84.5}&{81.15}\\
&ProGrad&{72.68}&{95.8}&92.13&81.46&94.87&87.01&40.39&75.0&65.92&84.38&81.59&79.2\\
&KgCoOp&71.2&95.03&93.23&74.87&92.9&87.03&36.27&73.4&69.37&74.93&81.43&77.26\\
&CLIPFit&71.53&\textbf{96.13}&\textbf{93.5}&82.43&96.37&87.37&{45.47}&\textbf{75.67}&71.57&\textbf{90.13}&83.83&\textbf{81.27}\\
\bottomrule   
\end{tabular}

}
\end{table*}
\textbf{LayerNorm gradient.} We visualize the squared sum of gradient from each LayerNorm layer in the image encoder in Fig. \ref{fig:change_iteration} (a). As observed, the magnitude of gradient in the first LayerNorm layer is much bigger than other layers. So the difference in change may be caused by the difference in gradient.

\textbf{Change in each iteration.} We visualize the change in first-layer text bias terms, first-layer 
 LayerNorm gain,  and first-layer LayerNorm bias for each iteration in Fig. \ref{fig:change_iteration} (b).  As observed, the change will increase smoothly and converge to some values.

\textbf{Analyses on other datasets.} We also conducted analyses on other datasets. The results for the EuroSAT dataset are shown in Fig. \ref{fig:change_1}, Fig. \ref{fig:change_loss_2}, and Fig. \ref{fig:change_loss_3}. The phenomena in the EuroSAT dataset are very similar to the DTD dataset.

\begin{table*}
\caption{Comparison with existing methods in the base-to-new generalization based on the \textbf{4-shot} settings. H: Harmonic mean.}
\label{tab:vit-4}
\centering
\small
\resizebox{0.9\linewidth}{!}{
\begin{tabular}{cc|ccccc|cc}
\toprule
              Datasets&metric& {CoOp}  &
              CLIP-adapter& {CoCoOp}                                              & {ProGrad}                                               & {KgCoOp}   &CLIPFit                                              \\
\midrule
\multirow{3}{*}{ImageNet} &Base&73.6 &74.23& 75.46 & 74.24 & 74.87  & 75.03
\\
&New&63.29 &67.93& 69.58 & 65.47 & 69.09 & 69.87
\\
&H&68.06 & 70.94& 72.4 & 69.58 & 71.86 & 72.36
\\
\midrule
\multirow{3}{*}{Caltech101} &Base&97.27 &97.23& 97.25 & 97.37 & 97.53 & 97.57

\\
&New&93.01 & 94.17 & 94.9 & 93.92 & 94.43  & 94.23\\
&H&95.09 &95.68& 96.06 & 95.61 & 95.95  & 95.87\\
\midrule
\multirow{3}{*}{OxfordPets} &Base&93.33 &93.8 & 94.59 & 94.08 & 94.68  & 94.93
\\
&New&95.69 &97.0 & 96.75 & 97.63 & 97.58  & 96.97\\
&H&94.5 &95.37& 95.66 & 95.82 & 96.11  & 95.94\\
\midrule
\multirow{3}{*}{StandfordCars} &Base&70.92 &69.43& 67.71 & 72.69 & 69.25  & 73.77
\\
&New&69.38 &73.0& 75.37 & 69.88 & 74.98 & 73.77\\
&H&70.14 & 71.17& 71.33 & 71.26 & 72.0  & 73.77
\\
\midrule
\multirow{3}{*}{Flowers} &Base&92.5 &87.93 & 84.75 & 92.46 & 91.3  & 91.03
\\
&New&70.12 &71.9 & 73.85 & 72.69 & 75.34 & 74.47
\\
&H&79.77 &79.11& 78.93 & 81.39 & 82.56  & 81.92\\
\midrule
\multirow{3}{*}{Food101} &Base&86.79 &90.2 & 89.79 & 88.91 & 90.3  & 90.2\\
&New&89.06 &90.97 & 90.99 & 90.18 & 91.39  & 91.23\\
&H&87.91 &90.58& 90.39 & 89.54 & 90.84  & 90.71\\
\midrule
\multirow{3}{*}{FGVCAircraft} &Base&33.21 &32.43 & 32.07 & 33.73 & 34.21 & 34.53\\
&New&28.57 &33.77& 33.93 & 30.09 & 32.81 & 31.47\\
&H&30.72 & 33.09& 32.97 & 31.81 & 33.5 & 32.93\\
\midrule
\multirow{3}{*}{Sun397} &Base&76.49 &77.7& 77.57 & 77.72 & 78.87  & 79.5
\\
&New&64.56 &75.67 & 76.96 & 71.93 & 75.64  & 77.77\\
&H&70.02 &76.67& 77.26 & 74.71 & 77.22  & 78.63\\
\midrule
\multirow{3}{*}{DTD} &Base&71.26 &67.43 & 67.44 & 71.06 & 73.65  & 74.37\\
&New&50.93 &55.43& 56.0 & 52.58 & 57.21  & 64.1\\
&H&59.4 &60.84& 61.19 & 60.44 & 64.4  & 68.85\\
\midrule
\multirow{3}{*}{EuroSAT} &Base&82.56 &81.9& 79.27 & 82.48 & 82.63 & 88.57\\
&New&53.04 &59.67& 65.44 & 56.43 & 59.98  & 76.7\\
&H&64.59 &69.04& 71.69 & 67.01 & 69.51  & 82.21\\
\midrule
\multirow{3}{*}{UCF101} &Base&79.97 &80.4 & 78.01 & 81.3 & 80.8 & 82.77\\
&New&65.98 &76.17 & 73.07 & 76.02 & 75.77  & 76.43\\
&H&72.3 &78.23& 75.46 & 78.57 & 78.2  & 79.47\\
\midrule
\multirow{3}{*}{AVG} &Base&78.43 &77.52& 76.72 & 79.18 & 78.92  & 80.21
\\
&New&68.03 &72.33& 73.35 & 71.14 & 73.11  & 75.18\\
&H&72.44 & 74.84& 74.85 & 74.62 & 75.9 & 77.61\\
\bottomrule               
\end{tabular}
}
\end{table*}

\begin{table*}
\caption{Comparison with existing methods in the base-to-new generalization based on the \textbf{8-shot} settings. H: Harmonic mean.}
\label{tab:vit-8}
\centering
\small
\resizebox{0.9\linewidth}{!}{
\begin{tabular}{cc|ccccc|cc}
\toprule
              Datasets&metric& {CoOp}  &
              CLIP-adapter & {CoCoOp}                                                & {ProGrad}                                                 & {KgCoOp}   &CLIPFit                                              \\
\midrule
\multirow{3}{*}{ImageNet} &Base&75.22 &75.07& 75.52 & 75.72 & 75.84 & 75.73
\\
&New&65.91 &67.6& 70.28 & 66.76 & 69.33  & 70.07\\
&H&70.26 &71.14& 72.81 & 70.96 & 72.44  & 72.79\\
\midrule
\multirow{3}{*}{Caltech101} &Base&97.81 &97.3& 97.76 & 98.0 & 97.68  & 97.83\\
&New&92.58 & 93.83& 93.63 & 93.38 & 94.1  & 93.8\\
&H&95.12 &95.53& 95.65 & 95.63 & 95.86  & 95.77\\
\midrule
\multirow{3}{*}{OxfordPets} &Base&94.19 &94.33& 95.5 & 94.47 & 94.81  & 94.83\\
&New&96.11 &96.83& 97.69 & 97.03 & 97.58  & 97.03\\
&H&95.14 &95.56& 96.58 & 95.73 & 96.18  & 95.92\\
\midrule
\multirow{3}{*}{StandfordCars} &Base&73.2 &72.13& 69.7 & 75.08 & 69.66  & 76.63\\
&New&67.44 &71.37 & 74.13 & 70.63 & 75.4  & 74.23\\
&H&70.2 &71.75& 71.85 & 72.79 & 72.42  & 75.41\\
\midrule
\multirow{3}{*}{Flowers} &Base&96.17 &94.27& 92.24 & 93.8&87.72 &  94.17\\
&New&69.41 & 70.67& 72.77 & 72.2&74.75  & 74.47\\
&H&80.63 &80.78& 81.36 & 81.59 &80.72 & 83.17\\
\midrule
\multirow{3}{*}{Food101} &Base&87.27 &90.27& 89.6 & 89.48 & 90.46  &90.33\\
&New&86.96 &90.7& 90.79 & 89.9 & 91.63  & 91.5\\
&H&87.11 &90.48& 90.19 & 89.69 & 91.04   & 90.91\\
\midrule
\multirow{3}{*}{FGVCAircraft} &Base&37.01 &35.47& 33.71 & 36.89 & 34.53 & 38.9\\
&New&38.45 &33.03& 32.15 & 31.67 & 34.95  & 32.43\\
&H&37.72 & 34.2& 32.91 & 34.08 & 34.74  & 35.37\\
\midrule
\multirow{3}{*}{Sun397} &Base&78.61 &79.53& 78.05 & 79.21 & 79.37  & 80.57\\
&New&66.25 &74.9 & 76.29 & 70.77 & 76.85  & 77.77\\
&H&71.9 &77.1& 77.16 & 74.75 & 78.09 & 79.15\\
\midrule
\multirow{3}{*}{DTD} &Base&76.97 &74.43& 73.03 & 74.42 & 69.72  & 77.87
\\
&New&51.81 &52.77 & 57.24 & 52.38 & 56.44  & 62.63\\
&H&61.93 &61.75& 64.18 & 61.48 & 62.38  & 69.42\\
\midrule
\multirow{3}{*}{EuroSAT} &Base&83.27 &80.23& 78.68 & 82.27 & 81.07  & 90.3\\
&New&50.59 &59.87& 56.03 & 58.52 & 63.13 & 73.0\\
&H&62.94 &68.57& 65.45 & 68.39 & 70.98 & 80.73\\
\midrule
\multirow{3}{*}{UCF101} &Base&82.85 &82.83& 80.4 & 82.61 & 81.16  & 84.4\\
&New&64.32 &74.53& 71.68 & 73.75 & 78.65  & 77.57\\
&H&72.42 &78.46& 75.79 & 77.93 & 79.89 & 80.84\\
\midrule
\multirow{3}{*}{AVG} &Base&80.74 &79.62& 78.56 & 80.62 & 78.37 & 81.96\\
&New&68.39 & 71.46 & 72.06 & 71.02  &73.75 & 74.95\\
&H&73.51 &75.32& 74.9 & 75.21 & 76.06  & 78.3\\
\bottomrule               
\end{tabular}}
\end{table*}

\end{document}